\documentclass{article}
\usepackage{times}
\usepackage{graphicx}
\usepackage{caption}
\usepackage{subcaption}
\usepackage{natbib}
\usepackage{algorithm}
\usepackage{algorithmic}
\usepackage{model_evolution}
\usepackage{placeins}
\usepackage{amsmath}
\usepackage{colortbl}
\usepackage{enumitem}

\usepackage[nohyperref, accepted]{icml2017}

\icmltitlerunning{Large-Scale Evolution}

\renewcommand{\cite}[1]{\citep{#1}}
\newcommand{\citing}[1]{#1}

\newcommand{\meanacc}{94.1}
\newcommand{\stddevacc}{0.4}
\newcommand{\meanflopsbase}{9}
\newcommand{\meanflopsexp}{19}
\newcommand{\bestacc}{94.6}
\newcommand{\totalflopsbase}{4}
\newcommand{\totalflopsexp}{20}
\newcommand{\bestparams}{5.4 M}
\newcommand{\controlacc}{87.3}
\newcommand{\controlflopsbase}{2}
\newcommand{\controlflopsexp}{17}
\newcommand{\darwinacc}{92.2}
\newcommand{\darwinflopsbase}{9}
\newcommand{\darwinflopsexp}{19}
\newcommand{\onehundredacc}{77.0}
\newcommand{\onehundredflopsbase}{2}
\newcommand{\onehundredflopsexp}{20}
\newcommand{\onehundredparams}{40.4 M}
\newcommand{\ensembleparams}{ensemb.}
\newcommand{\ensembleacc}{95.6}

\begin{document} 

\twocolumn[
\icmltitle{Large-Scale Evolution of Image Classifiers}

\begin{icmlauthorlist}
\icmlauthor{Esteban Real}{google_brain}
\icmlauthor{Sherry Moore}{google_brain}
\icmlauthor{Andrew Selle}{google_brain}
\icmlauthor{Saurabh Saxena}{google_brain}\\
\icmlauthor{Yutaka Leon Suematsu}{google_research}
\icmlauthor{Jie Tan}{google_brain}
\icmlauthor{Quoc V. Le}{google_brain} 
\icmlauthor{Alexey Kurakin}{google_brain}
\end{icmlauthorlist}

\icmlaffiliation{google_brain}{Google Brain, Mountain View, California, USA}
\icmlaffiliation{google_research}{Google Research, Mountain View, California, USA}

\icmlcorrespondingauthor{Esteban Real}{ereal@google.com}

\icmlkeywords{evolutionary algorithms, neural networks, neuro-evolution, neuroevolution, evolution, genetic algorithms, image classification}

\vskip 0.3in
]

\printAffiliationsAndNotice{}

\setcounter{footnote}{1}

\begin{abstract} 

Neural networks have proven effective at solving difficult problems but designing their architectures can be challenging, even for image classification problems alone. Our goal is to minimize human participation, so we employ evolutionary algorithms to discover such networks automatically. Despite significant computational requirements, we show that it is now possible to evolve models with accuracies within the range of those published in the last year. Specifically, we employ simple evolutionary techniques at unprecedented scales to discover models for the \mbox{CIFAR-10} and \mbox{CIFAR-100} datasets, starting from trivial initial conditions and reaching accuracies of \plainresult{\bestacc} (\plainresult{\ensembleacc} for ensemble) and \plainresult{\onehundredacc}, respectively. To do this, we use novel and intuitive mutation operators that navigate large search spaces; we stress that no human participation is required once evolution starts and that the output is a fully-trained model. Throughout this work, we place special emphasis on the repeatability of results, the variability in the outcomes and the computational requirements.
\end{abstract}

\section{Introduction}

\begin{table*}
\caption{Comparison with single-model hand-designed architectures. The ``C10+'' and ``C100+'' columns indicate the test accuracy on the data-augmented CIFAR-10 and CIFAR-100 datasets, respectively. The ``Reachable?'' column denotes whether the given hand-designed model lies within our search space. An entry of ``--'' indicates that no value was reported. The \textsuperscript{\textdagger} indicates a result reported by \citet{huang2016deep} instead of the original author. Much of this table was based on that presented in \citet{huang2016densely}.}
\label{hand_design_table}
\vskip 0.1in  
\begin{center}
\begin{small}
\begin{sc}
\begin{tabular}{lrrrc}
\hline
\abovespace\belowspace
Study & Params. & C10+ & C100+ & Reachable? \\
\hline
\abovespace
Maxout \cite{goodfellow2013maxout} & --~~~ & 90.7\% & 61.4\% & No \\
Network in Network \cite{lin2013network} & --~~~ & 91.2\% & --~~~ & No \\
All-CNN \cite{springenberg2014striving} & 1.3 M & 92.8\% & 66.3\% & Yes \\
Deeply Supervised \cite{lee2015deeply} & --~~~ & 92.0\% & 65.4\% & No \\
Highway \cite{srivastava2015highway} & 2.3 M & 92.3\% & 67.6\% & No \\
ResNet \cite{he2016deep} & 1.7 M & 93.4\% & 72.8\%\textsuperscript{\textdagger} & Yes \\
\rowcolor{Gray}
Evolution (ours) & \parbox[c][0.8cm]{1cm}{\hfill \bestparams \\ \hfill \onehundredparams} & \parbox{1cm}{\hfill \bestacc\% \\ } & \parbox{1cm}{\vspace{0.3cm} \hfill \onehundredacc\%} & N/A \\
Wide ResNet 28-10 \cite{zagoruyko2016wide} & 36.5 M & 96.0\% & 80.0\% & Yes \\
Wide ResNet 40-10+d/o \cite{zagoruyko2016wide} & 50.7 M & 96.2\% & 81.7\% & No \\
\belowspace
DenseNet \cite{huang2016densely} & 25.6 M & 96.7\% & 82.8\% & No \\
\end{tabular}
\end{sc}
\end{small}
\end{center}
\vskip -0.1in
\end{table*}

Neural networks can successfully perform difficult tasks where large amounts of training data are available \cite{he2015delving, weyand2016planet, silver2016mastering,wu2016google}. Discovering neural network architectures, however, remains a laborious task. Even within the specific problem of image classification, the state of the art was attained through many years of focused investigation by hundreds of researchers \citing{(\citet{krizhevsky2012imagenet, simonyan2014very, szegedy2015going, he2016deep, huang2016densely}, among many others)}. It is therefore not surprising that in recent years, techniques to automatically discover these architectures have been gaining popularity \cite{bergstra2012random, snoek2012practical, han2015learning, baker2016designing, zoph2016neural}. One of the earliest such ``neuro-discovery'' methods was {\em neuro-evolution} \cite{miller1989designing, stanley2002evolving, stanley2007compositional, bayer2009evolving, stanley2009hypercube, breuel2010automlp, pugh2013evolving, kim2015deep, zaremba2015empirical, fernando2016convolution, morse2016simple}. Despite the promising results, the deep learning community generally perceives evolutionary algorithms to be incapable of matching the accuracies of hand-designed models \cite{verbancsics2013generative, baker2016designing, zoph2016neural}. In this paper, we show that it is possible to evolve such competitive models today, given enough computational power.

We used slightly-modified known evolutionary algorithms and scaled up the computation to unprecedented levels, as far as we know. This, together with a set of novel and intuitive mutation operators, allowed us to reach competitive accuracies on the CIFAR-10 dataset. This dataset was chosen because it requires large networks to reach high accuracies, thus presenting a computational challenge. We also took a small first step toward generalization and evolved networks on the CIFAR-100 dataset. In transitioning from CIFAR-10 to CIFAR-100, we did not modify any aspect or parameter of our algorithm. Our typical neuro-evolution outcome on CIFAR-10 had a test accuracy with  \aggregateresult{\meanacc}{\stddevacc}{\meanflopsbase}{\meanflopsexp}, and our top model (by validation accuracy) had a test accuracy of \singleresult{\bestacc}{\totalflopsbase}{\totalflopsexp}. Ensembling the validation-top 2 models from each population reaches a test accuracy of \plainresult{\ensembleacc}, at no additional training cost. On CIFAR-100, our single experiment resulted in a test accuracy of \singleresult{\onehundredacc}{\onehundredflopsbase}{\onehundredflopsexp}. As far as we know, these are the most accurate results obtained on these datasets by automated discovery methods that start from trivial initial conditions.

Throughout this study, we placed special emphasis on the simplicity of the algorithm. In particular, it is a ``one-shot'' technique, producing a fully trained neural network requiring no post-processing. It also has few impactful meta-parameters (\ie parameters not optimized by the algorithm). Starting out with poor-performing models with no convolutions, the algorithm must evolve complex convolutional neural networks while navigating a fairly unrestricted search space: no fixed depth, arbitrary skip connections, and numerical parameters that have few restrictions on the values they can take. We also paid close attention to result reporting. Namely, we present the variability in our results in addition to the top value, we account for researcher degrees of freedom \cite{simmons2011false}, we study the dependence on the meta-parameters, and we disclose the amount of computation necessary to reach the main results. We are hopeful that our explicit discussion of computation cost could spark more study of efficient model search and training. Studying model performance normalized by computational investment allows consideration of economic concepts like opportunity cost.

\section{Related Work}

\begin{table*}
\caption{Comparison with automatically discovered architectures. The ``C10+'' and ``C100+'' contain the test accuracy on the data-augmented CIFAR-10 and CIFAR-100 datasets, respectively. An entry of ``--'' indicates that the information was not reported or is not known to us. For \citet{zoph2016neural}, we quote the result with the most similar search space to ours, as well as their best result. Please refer to Table~\ref{hand_design_table} for hand-designed results, including the state of the art. ``Discrete params.'' means that the parameters can be picked from a handful of values only (\eg strides $\in\{1,2,4\}$).}
\label{automated_design_table}
\vskip 0.1in  
\begin{center}
\begin{small}
\begin{sc}
\begin{tabular}{P{2.0 cm}P{2.7 cm}P{3.8 cm}P{2.8 cm}rrr}
\hline
\abovespace\belowspace
Study & Starting Point & Constraints & Post-Processing & Params. & C10+ & C100+ \\
\hline
\abovespace
Bayesian \cite{snoek2012practical} & 3 layers & fixed architecture, no skips & none & --~~~ & 90.5\% & --~~~ \\
\rule{0pt}{3ex}Q-Learning \cite{baker2016designing} & ~~~-- & discrete params., max. num. layers, no skips & tune, retrain & 11.2 M & 93.1\% & 72.9\% \\
\rule{0pt}{3ex}RL \cite{zoph2016neural} & 20 layers, 50\% skips & discrete params., exactly 20 layers & small grid search, retrain & 2.5 M & 94.0\% & --~~~ \\
\rule{0pt}{3ex}RL \cite{zoph2016neural} & 39 layers, 2 pool layers at 13 and 26, 50\% skips & discrete params., exactly 39 layers, 2 pool layers at 13 and 26 & add more filters, small grid search, retrain & 37.0 M & 96.4\% & --~~~ \\
\rowcolor{Gray}
\rule{0pt}{3ex}Evolution (ours) & single layer, zero convs. & power-of-2 strides & None & \parbox{1cm}{\vspace{0.2cm} \bestparams \\ \onehundredparams \\ \ensembleparams} & \parbox{1cm}{\vspace{0.2cm} \hfill \bestacc\% \\ \\ \parbox{1cm}{\hfill \ensembleacc\%} } & \parbox{1cm}{\vspace{0.2cm} \hfill \onehundredacc\%} \\
\end{tabular}
\end{sc}
\end{small}
\end{center}
\vskip -0.1in
\end{table*}

Neuro-evolution dates back many years \cite{miller1989designing}, originally being used only to evolve the weights of a fixed architecture. \citet{stanley2002evolving} showed that it was advantageous to simultaneously evolve the architecture using the {\em NEAT algorithm}. NEAT has three kinds of mutations: (i) modify a weight, (ii) add a connection between existing nodes, or (iii) insert a node while splitting an existing connection. It also has a mechanism for {\em recombining} two models into one and a strategy to promote diversity known as {\em fitness sharing} \cite{goldberg1987genetic}. Evolutionary algorithms represent the models using an encoding that is convenient for their purpose---analogous to nature's DNA. NEAT uses a {\em direct encoding}: every node and every connection is stored in the DNA. The alternative paradigm, {\em indirect encoding}, has been the subject of much neuro-evolution research \cite{gruau1993genetic, stanley2009hypercube, pugh2013evolving, kim2015deep, fernando2016convolution}. For example, the {\em CPPN} \cite{stanley2007compositional, stanley2009hypercube} allows for the evolution of repeating features at different scales. Also, \citet{kim2015deep} use an indirect encoding to improve the convolution filters in an initially highly-optimized fixed architecture.

Research on weight evolution is still ongoing \cite{morse2016simple} but the broader machine learning community defaults to back-propagation for optimizing neural network weights \cite{rumelhart1988learning}. Back-propagation and evolution can be combined as in \citet{stanley2009hypercube}, where only the structure is evolved. Their algorithm follows an alternation of architectural mutations and weight back-propagation. Similarly, \citet{breuel2010automlp} use this approach for hyper-parameter search. \citet{fernando2016convolution} also use back-propagation, allowing the trained weights to be {\em inherited} through the structural modifications.

The above studies create neural networks that are small in comparison to the typical modern architectures used for image classification \cite{he2016deep, huang2016densely}. Their focus is on the encoding or the efficiency of the evolutionary process, but not on the scale. When it comes to images, some neuro-evolution results reach the computational scale required to succeed on the MNIST dataset \cite{lecun1998mnist}. Yet, modern classifiers are often tested on realistic images, such as those in the CIFAR datasets \cite{krizhevsky2009learning}, which are much more challenging. These datasets require large models to achieve high accuracy.

Non-evolutionary neuro-discovery methods have been more successful at tackling realistic image data. \citet{snoek2012practical} used Bayesian optimization to tune 9 hyper-parameters for a fixed-depth architecture, reaching a new state of the art at the time. \citet{zoph2016neural} used reinforcement learning on a deeper fixed-length architecture. In their approach, a neural network---the ``discoverer''---constructs a convolutional neural network---the ``discovered''---one layer at a time. In addition to tuning layer parameters, they add and remove skip connections. This, together with some manual post-processing, gets them very close to the (current) state of the art. (Additionally, they surpassed the state of the art on a sequence-to-sequence problem.) \citet{baker2016designing} use Q-learning to also discover a network one layer at a time, but in their approach, the number of layers is decided by the discoverer. This is a desirable feature, as it would allow a system to construct shallow or deep solutions, as may be the requirements of the dataset at hand. Different datasets would not require specially tuning the algorithm. Comparisons among these methods are difficult because they explore very different search spaces and have very different initial conditions (Table~\ref{automated_design_table}).

Tangentially, there has also been neuro-evolution work on LSTM structure \cite{bayer2009evolving, zaremba2015empirical}, but this is beyond the scope of this paper. Also related to this work is that of \citet{saxena2016convolutional}, who embed convolutions with different parameters into a species of ``super-network'' with many parallel paths. Their algorithm then selects and ensembles paths in the super-network. Finally, canonical approaches to hyper-parameter search are {\em grid search} \citing{(used in \citet{zagoruyko2016wide}, for example)} and {\em random search}, the latter being the better of the two \cite{bergstra2012random}.

Our approach builds on previous work, with some important differences. We explore large model-architecture search spaces starting with basic initial conditions to avoid priming the system with information about known good strategies for the specific dataset at hand. Our encoding is different from the neuro-evolution methods mentioned above: we use a simplified graph as our DNA, which is transformed to a full neural network graph for training and evaluation (Section~\ref{methods_section}). Some of the mutations acting on this DNA are reminiscent of NEAT. However, instead of single nodes, one mutation can insert whole {\em layers}---\ie tens to hundreds of nodes at a time. We also allow for these layers to be removed, so that the evolutionary process can simplify an architecture in addition to complexifying it. Layer parameters are also mutable, but we do not prescribe a small set of possible values to choose from, to allow for a larger search space. We do not use fitness sharing. We report additional results using recombination, but for the most part, we used mutation only. On the other hand, we do use back-propagation to optimize the weights, which can be inherited across mutations. Together with a learning rate mutation, this allows the exploration of the space of learning rate schedules, yielding fully trained models at the end of the evolutionary process (Section~\ref{methods_section}). Tables~\ref{hand_design_table}~and~\ref{automated_design_table} compare our approach with hand-designed architectures and with other neuro-discovery techniques, respectively.

\section{Methods}
\label{methods_section}

\subsection{Evolutionary Algorithm}
\label{evolutionary_algorithm_section}

To automatically search for high-performing neural network architectures, we evolve a {\em population} of models. Each model---or {\em individual}---is a trained architecture. The model's accuracy on a separate validation dataset is a measure of the individual's quality or {\em fitness}. During each evolutionary step, a computer---a {\em worker}---chooses two individuals at random from this population and compares their fitnesses. The worst of the pair is immediately removed from the population---it is {\em killed}. The best of the pair is selected to be a {\em parent}, that is, to undergo {\em reproduction}. By this we mean that the worker creates a copy of the parent and modifies this copy by applying a {\em mutation}, as described below. We will refer to this modified copy as the {\em child}. After the worker creates the child, it trains this child, evaluates it on the validation set, and puts it back into the population. The child then becomes {\em alive}---\ie free to act as a parent. Our scheme, therefore, uses repeated pairwise competitions of random individuals, which makes it an example of {\em tournament selection} \cite{goldberg1991comparative}. Using pairwise comparisons instead of whole population operations prevents workers from idling when they finish early. Code and more detail about the methods described below can be found in Supplementary Section~\ref{methods_supplementary_section}.

Using this strategy to search large spaces of complex image models requires considerable computation. To achieve scale, we developed a massively-parallel, lock-free infrastructure. Many workers operate asynchronously on different computers. They do not communicate directly with each other. Instead, they use a shared file-system, where the population is stored. The file-system contains directories that represent the individuals. Operations on these individuals, such as the killing of one, are represented as atomic renames on the directory\footnote{The use of the file-name string to contain key information about the individual was inspired by \citet{breuel2010automlp}, and it speeds up disk access enormously. In our case, the file name contains the {\em state} of the individual ({\em alive}, {\em dead}, {\em training}, \etc).}. Occasionally, a worker may concurrently modify the individual another worker is operating on. In this case, the affected worker simply gives up and tries again. The {\em population size} is 1000 individuals, unless otherwise stated. The number of workers is always $\frac{1}{4}$ of the population size. To allow for long run-times with a limited amount of space, dead individuals' directories are frequently garbage-collected.

\subsection{Encoding and Mutations}
\label{encoding_and_mutations_section}

Individual architectures are encoded as a graph that we refer to as the {\em DNA}. In this graph, the vertices represent \mbox{rank-3} tensors or {\em activations}. As is standard for a convolutional network, two of the dimensions of the tensor represent the spatial coordinates of the image and the third is a number of channels. Activation functions are applied at the vertices and can be either (i) batch-normalization \cite{ioffe2015batch} with rectified linear units ({\em ReLUs}) or (ii) plain linear units. The graph's edges represent identity connections or convolutions and contain the mutable numerical parameters defining the convolution's properties. When multiple edges are incident on a vertex, their spatial scales or numbers of channels may not coincide. However, the vertex must have a single size and number of channels for its activations. The inconsistent inputs must be resolved. Resolution is done by choosing one of the incoming edges as the primary one. We pick this primary edge to be the one that is not a skip connection. The activations coming from the non-primary edges are reshaped through zeroth-order interpolation in the case of the size and through truncation/padding in the case of the number of channels, as in \citet{he2016deep}. In addition to the graph, the learning-rate value is also stored in the DNA.

A child is similar but not identical to the parent because of the action of a mutation. In each reproduction event, the worker picks a mutation at random from a predetermined set. The set contains the following mutations:
\begin{itemize}[noitemsep,topsep=0pt,leftmargin=*]
    \item {\sc Alter-learning-rate} (sampling details below).
    \item {\sc Identity} (effectively means ``keep training'').
    \item {\sc Reset-weights} (sampled as in \citet{he2015delving}, for example).
    \item {\sc Insert-convolution} (inserts a convolution at a random location in the ``convolutional backbone'', as in Figure~\ref{example_experiment_figure}. The inserted convolution has $3 \times 3$ filters, strides of $1$ or $2$ at random, number of channels same as input. May apply batch-normalization and ReLU activation or none at random).
    \item {\sc Remove-convolution}.
    \item {\sc Alter-stride} (only powers of 2 are allowed).
    \item {\sc Alter-number-of-channels} (of random conv.).
    \item {\sc Filter-size} (horizontal or vertical at random, on random convolution, odd values only).
    \item {\sc Insert-one-to-one} (inserts a one-to-one/identity connection, analogous to insert-convolution mutation).
    \item {\sc Add-skip} (identity between random layers).
    \item {\sc Remove-skip} (removes random skip).
\end{itemize}
These specific mutations were chosen for their similarity to the actions that a human designer may take when improving an architecture. This may clear the way for hybrid evolutionary--hand-design methods in the future. The probabilities for the mutations were not tuned in any way.

A mutation that acts on a numerical parameter chooses the new value at random around the existing value. All sampling is from uniform distributions. For example, a mutation acting on a convolution with 10 output channels will result in a convolution having between 5 and 20 output channels (that is, half to twice the original value). All values within the range are possible. As a result, the models are not constrained to a number of filters that is known to work well. The same is true for all other parameters, yielding a ``dense'' search space. In the case of the strides, this applies to the log-base-2 of the value, to allow for activation shapes to match more easily\footnote{For integer DNA parameters, we actually store and mutate a floating-point value. This allows multiple small mutations to have a cumulative effect in spite of integer round-off.}. In principle, there is also no upper limit to any of the parameters. All model depths are attainable, for example. Up to hardware constraints, the search space is unbounded. The dense and unbounded nature of the parameters result in the exploration of a truly large set of possible architectures.

\subsection{Initial Conditions}

Every evolution {\em experiment} begins with a population of simple individuals, all with a learning rate of 0.1. They are all very bad performers. Each initial individual constitutes just a single-layer model with no convolutions. This conscious choice of poor initial conditions forces evolution to make the discoveries by itself. The experimenter contributes mostly through the choice of mutations that demarcate a search space. Altogether, the use of poor initial conditions and a large search space limits the experimenter's impact. In other words, it prevents the experimenter from ``rigging'' the experiment to succeed.

\subsection{Training and Validation}
\label{training_and_validation_section}

Training and validation is done on the CIFAR-10 dataset. This dataset consists of 50,000 training examples and 10,000 test examples, all of which are 32 x 32 color images labeled with 1 of 10 common object classes \cite{krizhevsky2009learning}. 5,000 of the training examples are held out in a validation set. The remaining 45,000 examples constitute our actual training set. The training set is augmented as in \citet{he2016deep}. The CIFAR-100 dataset has the same number of dimensions, colors and examples as CIFAR-10, but uses 100 classes, making it much more challenging.

Training is done with {\em TensorFlow} \cite{abadi2016tensorflow}, using SGD with a momentum of 0.9 \cite{sutskever2013importance}, a batch size of 50, and a weight decay of 0.0001. Each training runs for 25,600 steps, a value chosen to be brief enough so that each individual could be trained in a few seconds to a few hours, depending on model size. The loss function is the cross-entropy. Once training is complete, a single evaluation on the validation set provides the accuracy to use as the individual's fitness. Ensembling was done by majority voting during the testing evaluation. The models used in the ensemble were selected by validation accuracy.

\subsection{Computation cost}

To estimate computation costs, we identified the basic TensorFlow ({\em TF}) operations used by our model training and validation, like convolutions, generic matrix multiplications, \etc. For each of these TF operations, we estimated the theoretical number of floating-point operations ({\em FLOPs}) required. This resulted in a map from TF operation to FLOPs, which is valid for all our experiments.

For each individual within an evolution experiment, we compute the total FLOPs incurred by the TF operations in its architecture over one batch of examples, both during its training ($F_t$ FLOPs) and during its validation ($F_v$ FLOPs). Then we assign to the individual the cost $F_t N_t + F_v N_v$, where $N_t$ and $N_v$ are the number of training and validation batches, respectively. The cost of the experiment is then the sum of the costs of all its individuals.

We intend our FLOPs measurement as a coarse estimate only. We do not take into account input/output, data preprocessing, TF graph building or memory-copying operations. Some of these unaccounted operations take place once per training run or once per step and some have a component that is constant in the model size (such as disk-access latency or input data cropping). We therefore expect the estimate to be more useful for large architectures (for example, those with many convolutions).

\subsection{Weight Inheritance}
\label{weight_inheritance_section}

We need architectures that are trained to completion within an evolution experiment. If this does not happen, we are forced to retrain the best model at the end, possibly having to explore its hyper-parameters. Such extra exploration tends to depend on the details of the model being retrained. On the other hand, 25,600 steps are not enough to fully train each individual. Training a large model to completion is prohibitively slow for evolution. To resolve this dilemma, we allow the children to inherit the parents' weights whenever possible. Namely, if a layer has matching shapes, the weights are preserved. Consequently, some mutations preserve all the weights (like the identity or learning-rate mutations), some preserve none (the weight-resetting mutation), and most preserve some but not all. An example of the latter is the filter-size mutation: only the filters of the convolution being mutated will be discarded.

\subsection{Reporting Methodology}
\label{reporting_section}

To avoid over-fitting, neither the evolutionary algorithm nor the neural network training ever see the testing set. Each time we refer to ``the best model'', we mean the model with the highest validation accuracy. However, we always report the test accuracy. This applies not only to the choice of the best individual within an experiment, but also {\em to the choice of the best experiment}. Moreover, we only include experiments that we managed to reproduce, unless explicitly noted. Any statistical analysis was fully decided upon before seeing the results of the experiment reported, to avoid tailoring our analysis to our experimental data \cite{simmons2011false}.

\section{Experiments and Results}
\label{results_section}

\begin{figure*}
    \vskip 0.2in
    \begin{centering}
        \centerline{\includegraphics[width=\textwidth]{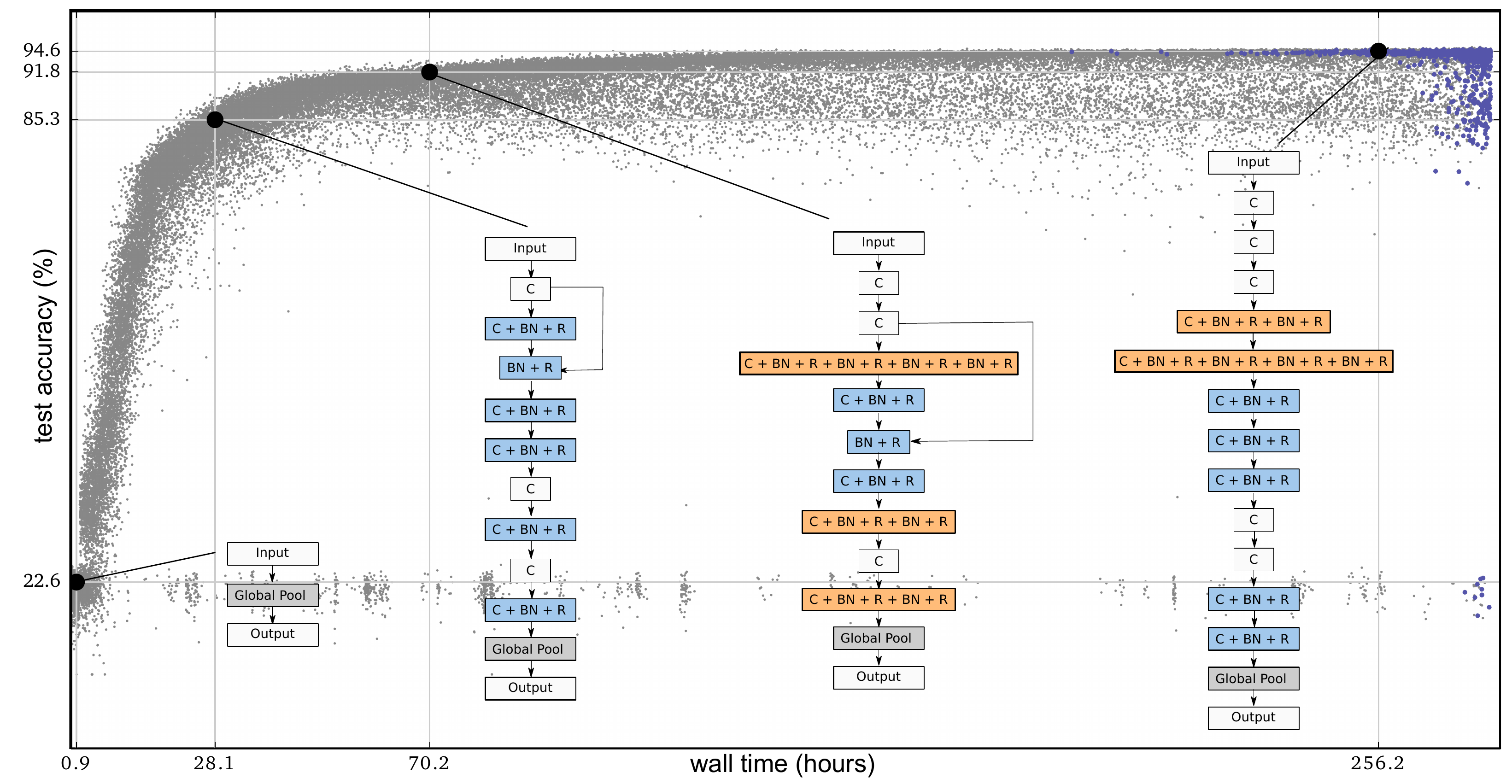}}
        \caption{Progress of an evolution experiment. Each dot represents an individual in the population. Blue dots (darker, top-right) are alive. The rest have been killed. The four diagrams show examples of discovered architectures. These correspond to the best individual (right-most) and three of its ancestors. The best individual was selected by its validation accuracy. Evolution sometimes stacks convolutions without any nonlinearity in between (``$C$'', white background), which are mathematically equivalent to a single linear operation. Unlike typical hand-designed architectures, some convolutions are followed by more than one nonlinear function (``$C+BN+R+BN+R+...$'', orange background).}
        \label{example_experiment_figure}
    \end{centering}
    \vskip 0.2in
\end{figure*}

We want to answer the following questions:
\begin{itemize}[noitemsep,topsep=0pt,leftmargin=*]
    \item Can a simple one-shot evolutionary process start from trivial initial conditions and yield fully trained models that rival hand-designed architectures?
    \item What are the variability in outcomes, the parallelizability, and the computation cost of the method?
    \item Can an algorithm designed iterating on CIFAR-10 be applied, without any changes at all, to CIFAR-100 and still produce competitive models?
\end{itemize}

We used the algorithm in Section~\ref{methods_section} to perform several experiments. Each experiment evolves a population in a few days, typified by the example in Figure~\ref{example_experiment_figure}. The figure also contains examples of the architectures discovered, which turn out to be surprisingly simple. Evolution attempts skip connections but frequently rejects them.

\begin{figure}
    \vskip 0.2in
    \begin{centering}
        \centerline{\includegraphics[width=\columnwidth]{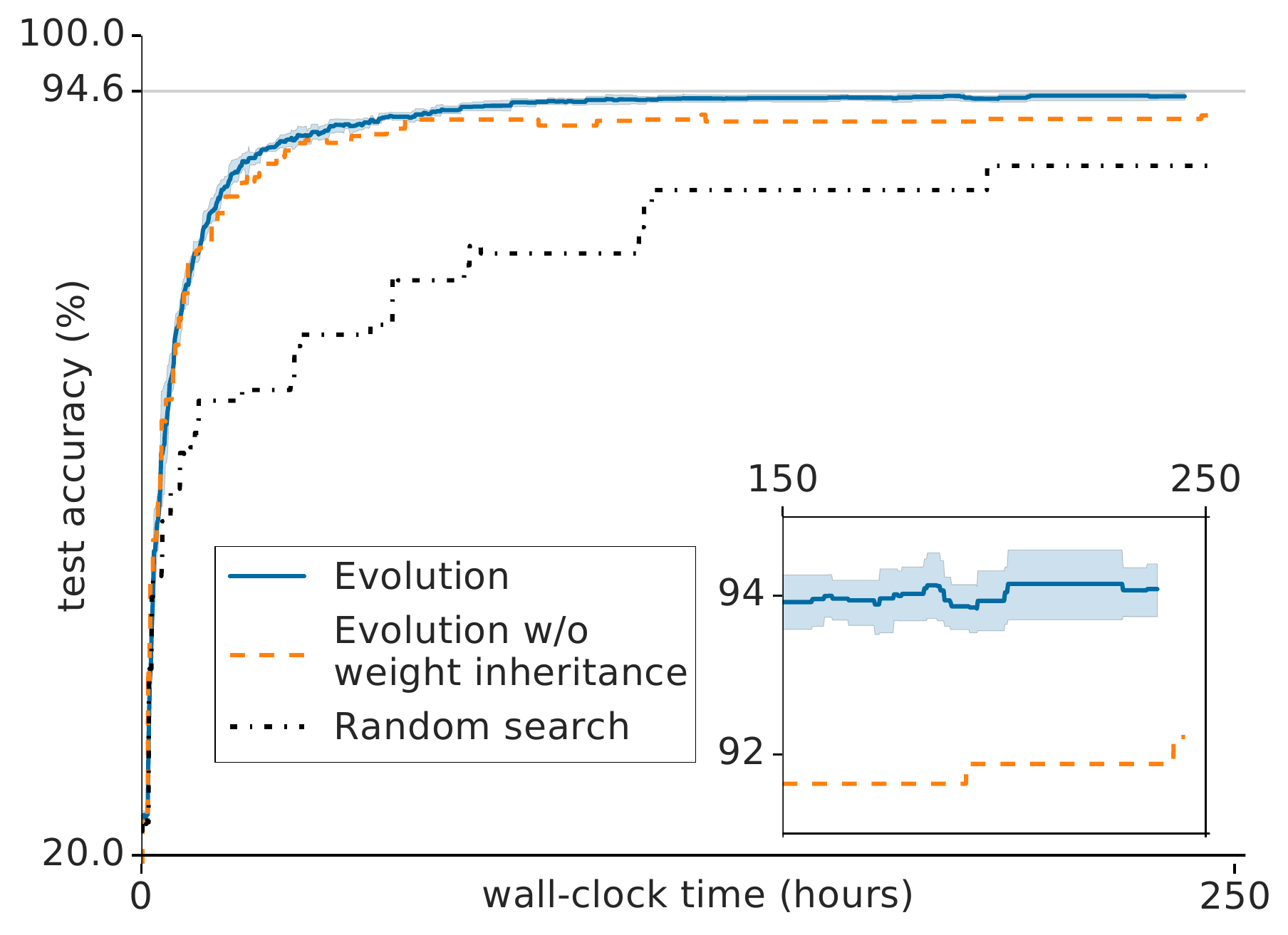}}
        \caption{Repeatability of results and controls. In this plot, the vertical axis at wall-time $t$ is defined as the test accuracy of the individual with the highest validation accuracy that became alive at or before $t$. The inset magnifies a portion of the main graph. The curves show the progress of various experiments, as follows. The top line (solid, blue) shows the mean test accuracy across 5 large-scale evolution experiments. The shaded area around this top line has a width of $\pm 2 \sigma$ (clearer in inset). The next line down (dashed, orange, main graph and inset) represents a single experiment in which weight-inheritance was disabled, so every individual has to train from random weights. The lowest curve (dotted-dashed) is a random-search control. All experiments occupied the same amount and type of hardware. A small amount of noise in the generalization from the validation to the test set explains why the lines are not monotonically increasing. Note the narrow width of the $\pm 2 \sigma$ area (main graph and inset), which shows that the high accuracies obtained in evolution experiments are repeatable.}
        \label{lamarckian_experiments_figure}
    \end{centering}
    \vskip 0.2in
\end{figure}

To get a sense of the variability in outcomes, we repeated the experiment 5 times. Across all 5 experiment runs, the best model by validation accuracy has a testing accuracy of \acc{\bestacc}. Not all experiments reach the same accuracy, but they get close ($\mu\!=\!\meanacc\%$, $\sigma\!=\!\stddevacc$). Fine differences in the experiment outcome may be somewhat distinguishable by validation accuracy ($\textrm{correlation coefficient}\!=\!0.894$). The total amount of computation {\em across all 5 experiments} was \flops{\totalflopsbase}{\totalflopsexp} (or \flops{\meanflopsbase}{\meanflopsexp} on average per experiment). Each experiment was distributed over 250 parallel workers (Section~\ref{evolutionary_algorithm_section}). Figure~\ref{lamarckian_experiments_figure} shows the progress of the experiments in detail.

As a control, we disabled the selection mechanism, thereby reproducing and killing random individuals. This is the form of random search that is most compatible with our infrastructure. The probability distributions for the parameters are implicitly determined by the mutations. This control only achieves an accuracy of \acc{\controlacc} in the same amount of run time on the same hardware (Figure~\ref{lamarckian_experiments_figure}). The total amount of computation was \flops{\controlflopsbase}{\controlflopsexp}. The low FLOP count is a consequence of random search generating many small, inadequate models that train quickly but consume roughly constant amounts of setup time (not included in the FLOP count). We attempted to minimize this overhead by avoiding unnecessary disk access operations, to no avail: too much overhead remains spent on a combination of neural network setup, data augmentation, and training step initialization.

We also ran a partial control where the weight-inheritance mechanism is disabled. This run also results in a lower accuracy (\acc{\darwinacc}) in the same amount of time (Figure~\ref{lamarckian_experiments_figure}), using \flops{\darwinflopsbase}{\darwinflopsexp}. This shows that weight inheritance is important in the process.

Finally, we applied our neuro-evolution algorithm, without any changes and with the same meta-parameters, to CIFAR-100. Our only experiment reached an accuracy of \acc{\onehundredacc}, using \flops{\onehundredflopsbase}{\onehundredflopsexp}. We did not attempt other datasets. Table~\ref{hand_design_table} shows that both the CIFAR-10 and CIFAR-100 results are competitive with modern hand-designed networks.

\section{Analysis}
\label{analysis_section}

\paragraph{Meta-parameters.} We observe that populations evolve until they plateau at some local optimum (Figure~\ref{lamarckian_experiments_figure}). The fitness (\ie validation accuracy) value at this optimum varies between experiments (Figure~\ref{lamarckian_experiments_figure}, inset). Since not all experiments reach the highest possible value, some populations are getting ``trapped'' at inferior local optima. This entrapment is affected by two important meta-parameters (\ie parameters that are not optimized by the algorithm). These are the population size and the number of training steps per individual. Below we discuss them and consider their relationship to local optima.

\begin{figure}
    \vskip 0.2in
    \begin{centering}
        \begin{subfigure}[b]{0.49\columnwidth}
            \begin{centering}
                \centerline{\includegraphics[width=\columnwidth]{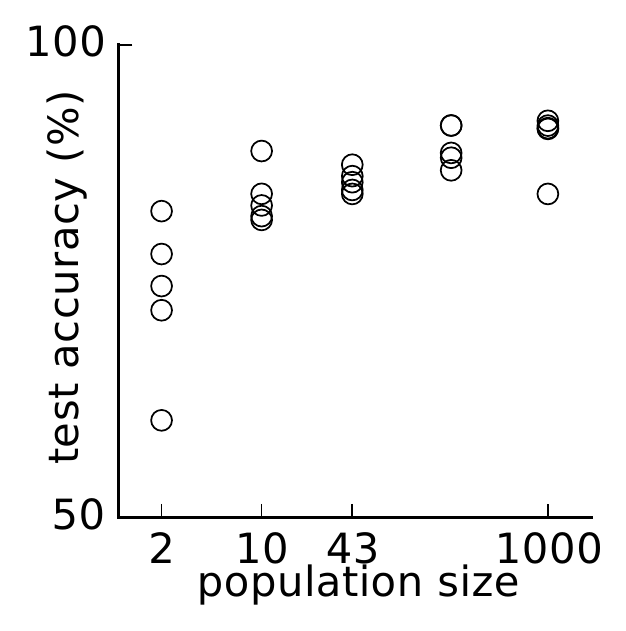}}
            \end{centering}
        \end{subfigure}
        \begin{subfigure}[b]{0.49\columnwidth}
            \begin{centering}
                \centerline{\includegraphics[width=\columnwidth]{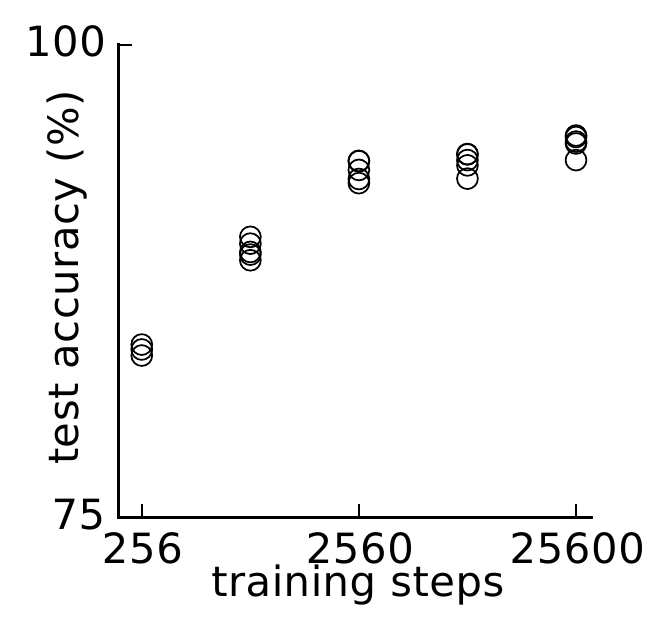}}
            \end{centering}
        \end{subfigure}
        \caption{Dependence on meta-parameters. In both graphs, each circle represents the result of a full evolution experiment. Both vertical axes show the test accuracy for the individual with the highest validation accuracy at the end of the experiment. All populations evolved for the same total wall-clock time. There are 5 data points at each horizontal axis value. LEFT: effect of population size. To economize resources, in these experiments the number of individual training steps is only $2560$. Note how the accuracy increases with population size. RIGHT: effect of number of training steps per individual. Note how the accuracy increases with more steps.}
        \label{meta_parameters_figure}
    \end{centering}
    \vskip 0.2in
\end{figure}

\vspace{-1em}
\paragraph{Effect of population size.} Larger populations explore the space of models more thoroughly, and this helps reach better optima (Figure~\ref{meta_parameters_figure}, left). Note, in particular, that a population of size $2$ can get trapped at very low fitness values. Some intuition about this can be gained by considering the fate of a {\em super-fit} individual, \ie an individual such that any one architectural mutation reduces its fitness (even though a sequence of many mutations may improve it). In the case of a population of size 2, if the super-fit individual wins once, it will win every time. After the first win, it will produce a child that is one mutation away. By definition of super-fit, therefore, this child is inferior\footnote{Except after identity or learning rate mutations, but these produce a child with the same architecture as the parent.}. Consequently, in the next round of tournament selection, the super-fit individual competes against its child and wins again. This cycle repeats forever and the population is trapped. Even if a sequence of two mutations would allow for an ``escape'' from the local optimum, such a sequence can never take place. This is only a rough argument to heuristically suggest why a population of size 2 is easily trapped. More generally, Figure~\ref{meta_parameters_figure} (left) empirically demonstrates a benefit from an increase in population size. Theoretical analyses of this dependence are quite complex and assume very specific models of population dynamics; often larger populations are better at handling local optima, at least beyond a size threshold (\citet{weinreich2005rapid} and references therein).

\vspace{-1em}
\paragraph{Effect of number of training steps.} The other meta-parameter is the number $T$ of training steps for each individual. Accuracy increases with $T$ (Figure~\ref{meta_parameters_figure}, right). Larger $T$ means an individual needs to undergo fewer identity mutations to reach a given level of training.

\vspace{-1em}
\paragraph{Escaping local optima.} While we might increase population size or number of steps to prevent a trapped population from forming, we can also free an already trapped population. For example, increasing the {\em mutation rate} or resetting all the weights of a population (Figure~\ref{escape_figure}) work well but are quite costly (more details in Supplementary Section~\ref{escaping_local_optima_supplementary_section}).

\begin{figure}
    \vskip 0.2in
    \begin{centering}
        \begin{subfigure}[b]{1.0\columnwidth}
            \begin{centering}
                \centerline{\includegraphics[width=\textwidth]{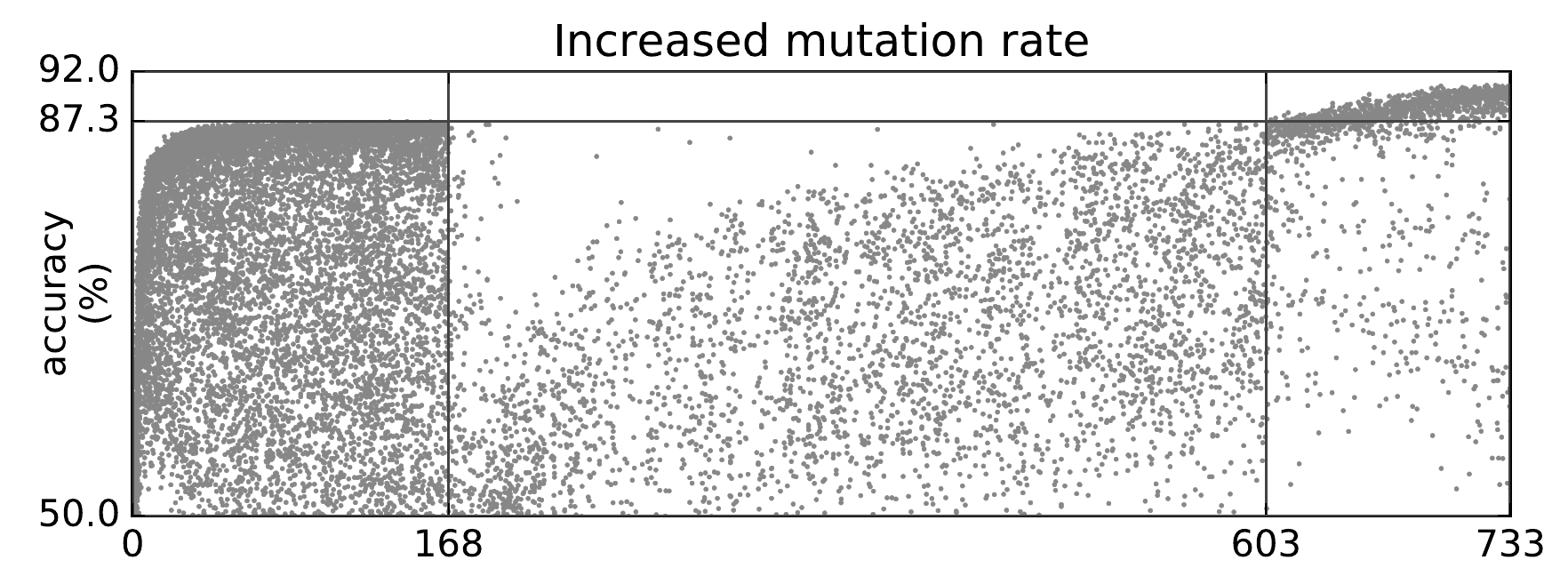}}
            \end{centering}
        \end{subfigure}
        \begin{subfigure}[b]{1.0\columnwidth}
            \begin{centering}
                \centerline{\includegraphics[width=\textwidth]{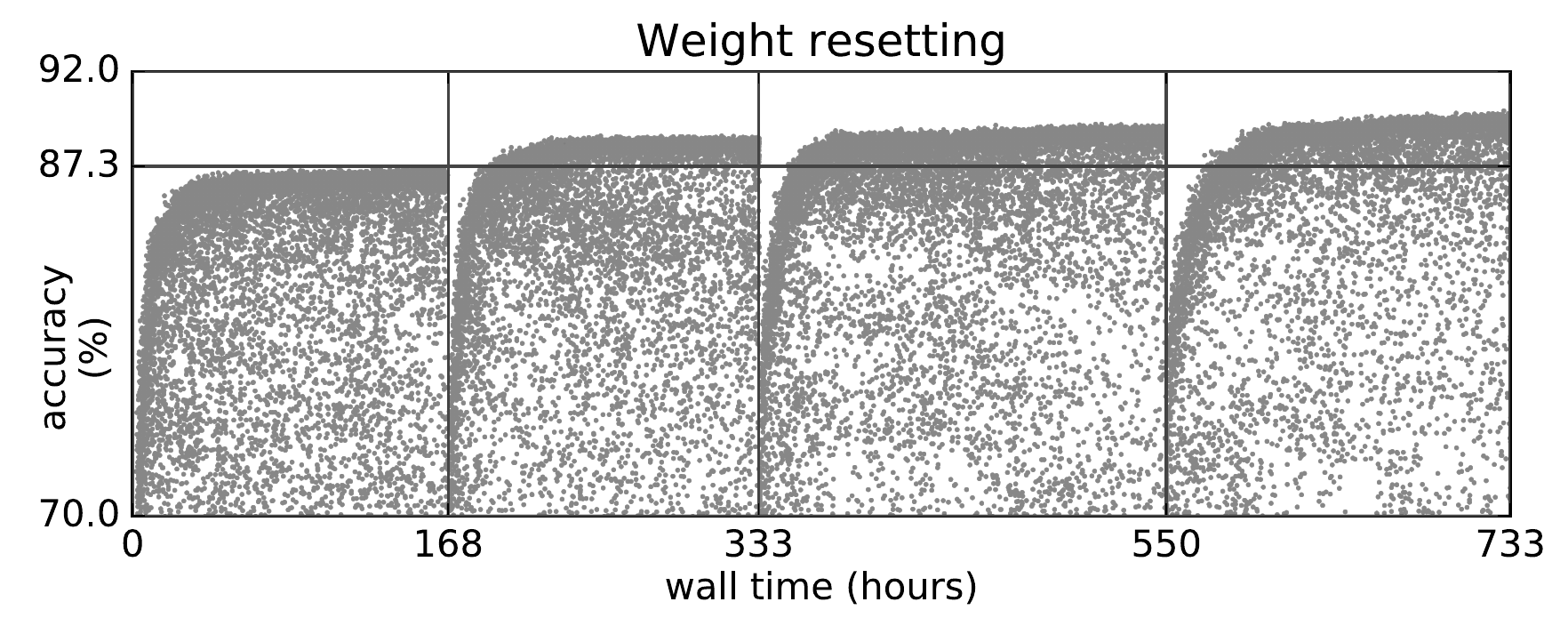}}
            \end{centering}
        \end{subfigure}
        \caption{Escaping local optima in two experiments. We used smaller populations and fewer training steps per individual ($2560$) to make it more likely for a population to get trapped and to reduce resource usage. Each dot represents an individual. The vertical axis is the accuracy. TOP: example of a population of size 100 escaping a local optimum by using a period of increased mutation rate in the middle (Section~\ref{analysis_section}). BOTTOM: example of a population of size 50 escaping a local optimum by means of three consecutive weight resetting events (Section~\ref{analysis_section}). Details in Supplementary Section~\ref{escaping_local_optima_supplementary_section}.}
        \label{escape_figure}
    \end{centering}
    \vskip 0.2in
\end{figure}

\vspace{-1em}
\paragraph{Recombination.} None of the results presented so far used recombination. However, we explored three forms of recombination in additional experiments. Following \citet{tuson1998adapting}, we attempted to evolve the mutation probability distribution too. On top of this, we employed a recombination strategy by which a child could inherit structure from one parent and mutation probabilities from another. The goal was to allow individuals that progressed well due to good mutation choices to quickly propagate such choices to others. In a separate experiment, we attempted recombining the trained weights from two parents in the hope that each parent may have learned different concepts from the training data. In a third experiment, we recombined structures so that the child fused the architectures of both parents side-by-side, generating wide models fast. While none of these approaches improved our recombination-free results, further study seems warranted.

\section{Conclusion}

In this paper we have shown that (i) neuro-evolution is capable of constructing large, accurate networks for two challenging and popular image classification benchmarks; (ii) neuro-evolution can do this starting from trivial initial conditions while searching a very large space; (iii) the process, once started, needs no experimenter participation; and (iv) the process yields fully trained models. Completely training models required weight inheritance (Sections~\ref{weight_inheritance_section}). In contrast to reinforcement learning, evolution provides a natural framework for weight inheritance: mutations can be constructed to guarantee a large degree of similarity between the original and mutated models---as we did. Evolution also has fewer tunable meta-parameters with a fairly predictable effect on the variance of the results, which can be made small.

While we did not focus on reducing computation costs, we hope that future algorithmic and hardware improvement will allow more economical implementation. In that case, evolution would become an appealing approach to neuro-discovery for reasons beyond the scope of this paper. For example, it ``hits the ground running'', improving on arbitrary initial models as soon as the experiment begins. The mutations used can implement recent advances in the field and can be introduced without having to restart an experiment. Furthermore, recombination can merge improvements developed by different individuals, even if they come from other populations. Moreover, it may be possible to combine neuro-evolution with other automatic architecture discovery methods.

\FloatBarrier

\section*{Acknowledgements} 
We wish to thank Vincent Vanhoucke, Megan Kacholia, Rajat Monga, and especially Jeff Dean for their support and valuable input; Geoffrey Hinton, Samy Bengio, Thomas Breuel, Mark DePristo, Vishy Tirumalashetty, Martin Abadi, Noam Shazeer, Yoram Singer, Dumitru Erhan, Pierre Sermanet, Xiaoqiang Zheng, Shan Carter and Vijay Vasudevan for helpful discussions; Thomas Breuel, Xin Pan and Andy Davis for coding contributions; and the larger Google Brain team for help with {\em TensorFlow} and training vision models.

\bibliography{model_evolution}
\bibliographystyle{icml2017}

\FloatBarrier
\clearpage
\renewcommand{\figurename}{Supplementary Figure}
\renewcommand{\tablename}{Supplementary Table}
\renewcommand{\thesection}{S\arabic{section}}
\setcounter{section}{0}
\setcounter{figure}{0}
\setcounter{table}{0}
\onecolumn
\icmltitle{Large-Scale Evolution of Image Classifiers}

{\LARGE Supplementary Material}

\section{Methods Details}

\label{methods_supplementary_section}

This section contains additional implementation details, roughly following the order in Section~\ref{methods_section}. Short code snippets illustrate the ideas. The code is not intended to run on its own and it has been highly edited for clarity.

In our implementation, each worker runs an outer loop that is responsible for selecting a pair of random individuals from the population. The individual with the highest fitness usually becomes a parent and the one with the lowest fitness is usually killed (Section \ref{evolutionary_algorithm_section}). Occasionally, either of these two actions is not carried out in order to keep the population size close to a set-point:
\smallskip
\begin{lstlisting}[language=Python]
def evolve_population(self):
  # Iterate indefinitely.
  while True:
    # Select two random individuals from the population.
    valid_individuals = []
    for individual in self.load_individuals():  # Only loads the IDs and states.
      if individual.state in [TRAINING, ALIVE]:
        valid_individuals.append(individual)
    individual_pair = random.sample(valid_individuals, 2)

    for individual in individual_pair:
      # Sync changes from other workers from file-system. Loads everything else.
      individual.update_if_necessary()

      # Ensure the individual is fully trained.
      if individual.state == TRAINING:
        self._train(individual)
        
    # Select by fitness (accuracy).
    individual_pair.sort(key=lambda i: i.fitness, reverse=True)
    better_individual = individual_pair[0]
    worse_individual = individual_pair[1]
    
    # If the population is not too small, kill the worst of the pair.
    if self._population_size() >= self._population_size_setpoint:
      self._kill_individual(worse_individual)

    # If the population is not too large, reproduce the best of the pair.
    if self._population_size() < self._population_size_setpoint:
      self._reproduce_and_train_individual(better_individual)
\end{lstlisting}

Much of the code is wrapped in try-except blocks to handle various kinds of errors. These have been removed from the code snippets for clarity. For example, the method above would be wrapped like this:
\smallskip
\begin{lstlisting}[language=Python]
def evolve_population(self):
  while True:
  try:
    # Select two random individuals from the population.
    ...
  except:
    except exceptions.PopulationTooSmallException:
      self._create_new_individual()
      continue
    except exceptions.ConcurrencyException:
      # Another worker did something that interfered with the action of this worker.
      # Abandon the current task and keep going.
      continue
\end{lstlisting}

The encoding for an individual is represented by a serializable \lstinline{DNA} class instance containing all information except for the trained weights (Section~\ref{encoding_and_mutations_section}). For all results in this paper, this encoding is a directed, acyclic graph where edges represent convolutions and vertices represent nonlinearities. This is a sketch of the \lstinline{DNA} class:
\smallskip
\begin{lstlisting}
class DNA(object):

  def __init__(self, dna_proto):
    """Initializes the `DNA` instance from a protocol buffer.
    
    The `dna_proto` is a protocol buffer used to restore the DNA state from disk.
    Together with the corresponding `to_proto` method, they allow for a
    serialization-deserialization mechanism.
    """
    # Allows evolving the learning rate, i.e. exploring the space of
    # learning rate schedules.
    self.learning_rate = dna_proto.learning_rate
    
    self._vertices = {}  # String vertex ID to `Vertex` instance.
    for vertex_id in dna_proto.vertices:
      vertices[vertex_id] = Vertex(vertex_proto=dna_sproto.vertices[vertex_id])

    self._edges = {}  # String edge ID to `Edge` instance.
    for edge_id in dna_proto.edges:
      mutable_edges[edge_id] = Edge(edge_proto=dna_proto.edges[edge_id])
    
    ...
    
  def to_proto(self):
    """Returns this instance in protocol buffer form."""
    dna_proto = dna_pb2.DnaProto(learning_rate=self.learning_rate)

    for vertex_id, vertex in self._vertices.iteritems():
      dna_proto.vertices[vertex_id].CopyFrom(vertex.to_proto())

    for edge_id, edge in self._edges.iteritems():
      dna_proto.edges[edge_id].CopyFrom(edge.to_proto())
      
    ...

    return dna_proto
    
  def add_edge(self, dna, from_vertex_id, to_vertex_id, edge_type, edge_id):
    """Adds an edge to the DNA graph, ensuring internal consistency."""
    # `EdgeProto` defines defaults for other attributes.
    edge = Edge(EdgeProto(
        from_vertex=from_vertex_id, to_vertex=to_vertex_id, type=edge_type))
    self._edges[edge_id] = edge
    self._vertices[from_vertex_id].edges_out.add(edge_id)
    self._vertices[to_vertex].edges_in.add(edge_id)
    return edge
    
  # Other methods like `add_edge` to manipulate the graph structure.
  ...
\end{lstlisting}
The \lstinline{DNA} holds \lstinline{Vertex} and \lstinline{Edge} instances. The \lstinline{Vertex} class looks like this:
\smallskip
\begin{lstlisting}
class Vertex(object):

  def __init__(self, vertex_proto):
    # Relationship to the rest of the graph.
    self.edges_in = set(vertex_proto.edges_in)  # Incoming edge IDs.
    self.edges_out = set(vertex_proto.edges_out)  # Outgoing edge IDs.

    # The type of activations.
    if vertex_proto.HasField('linear'):
      self.type = LINEAR  # Linear activations.
    elif vertex_proto.HasField('bn_relu'):
      self.type = BN_RELU  # ReLU activations with batch-normalization.
    else:
      raise NotImplementedError()
    
    # Some parts of the graph can be prevented from being acted upon by mutations.
    # The following boolean flags control this.
    self.inputs_mutable = vertex_proto.inputs_mutable
    self.outputs_mutable = vertex_proto.outputs_mutable
    self.properties_mutable = vertex_proto.properties_mutable
    
    # Each vertex represents a 2^s x 2^s x d block of nodes. s and d are positive
    # integers computed dynamically from the in-edges. s stands for "scale" so
    # that 2^x x 2^s is the spatial size of the activations. d stands for "depth",
    # the number of channels.
    
  def to_proto(self):
    ...
\end{lstlisting}
The \lstinline{Edge} class looks like this:
\smallskip
\begin{lstlisting}
class Edge(object):

  def __init__(self, edge_proto):
    # Relationship to the rest of the graph.
    self.from_vertex = edge_proto.from_vertex  # Source vertex ID.
    self.to_vertex = edge_proto.to_vertex  # Destination vertex ID.
    
    if edge_proto.HasField('conv'):
      # In this case, the edge represents a convolution.
      self.type = CONV
      
      # Controls the depth (i.e. number of channels) in the output, relative to the
      # input. For example if there is only one input edge with a depth of 16 channels
      # and `self._depth_factor` is 2, then this convolution will result in an output
      # depth of 32 channels. Multiple-inputs with conflicting depth must undergo
      # depth resolution first.
      self.depth_factor = edge_proto.conv.depth_factor
      
      # Control the shape of the convolution filters (i.e. transfer function).
      # This parameterization ensures that the filter width and height are odd
      # numbers: filter_width = 2 * filter_half_width + 1.
      self.filter_half_width = edge_proto.conv.filter_half_width
      self.filter_half_height = edge_proto.conv.filter_half_height
      
      # Controls the strides of the convolution. It will be 2^stride_scale.
      # Note that conflicting input scales must undergo scale resolution. This
      # controls the spatial scale of the output activations relative to the
      # spatial scale of the input activations.
      self.stride_scale = edge_proto.conv.stride_scale
    elif edge_spec.HasField('identity'):
      self.type = IDENTITY
    else:
      raise NotImplementedError()
      
    # In case depth or scale resolution is necessary due to conflicts in inputs,
    # These integer parameters determine which of the inputs takes precedence in
    # deciding the resolved depth or scale.
    self.depth_precedence = edge_proto.depth_precedence
    self.scale_precedence = edge_proto.scale_precedence

  def to_proto(self):
    ...
\end{lstlisting}

Mutations act on \lstinline{DNA} instances. The set of mutations restricts the space explored somewhat (Section~\ref{encoding_and_mutations_section}). The following are some example mutations. The \lstinline{AlterLearningRateMutation} simply randomly modifies the attribute in the \lstinline{DNA}:
\smallskip
\begin{lstlisting}
class AlterLearningRateMutation(Mutation):
  """Mutation that modifies the learning rate."""

  def mutate(self, dna):
    mutated_dna = copy.deepcopy(dna)
    
    # Mutate the learning rate by a random factor between 0.5 and 2.0,
    # uniformly distributed in log scale.
    factor = 2**random.uniform(-1.0, 1.0)
    mutated_dna.learning_rate = dna.learning_rate * factor
    
    return mutated_dna
\end{lstlisting}
Many mutations modify the structure. Mutations to insert and excise vertex-edge pairs build up a main convolutional column, while mutations to add and remove edges can handle the skip connections. For example, the \lstinline{AddEdgeMutation} can add a skip connection between random vertices.
\smallskip
\begin{lstlisting}
class AddEdgeMutation(Mutation):
  """Adds a single edge to the graph."""

  def mutate(self, dna):
    # Try the candidates in random order until one has the right connectivity.
    for from_vertex_id, to_vertex_id  in self._vertex_pair_candidates(dna):
      mutated_dna = copy.deepcopy(dna)
      if (self._mutate_structure(mutated_dna, from_vertex_id, to_vertex_id)):
        return mutated_dna
    raise exceptions.MutationException()  # Try another mutation.

  def _vertex_pair_candidates(self, dna):
    """Yields connectable vertex pairs."""
    from_vertex_ids = _find_allowed_vertices(dna, self._to_regex, ...)
    if not from_vertex_ids:
      raise exceptions.MutationException()  # Try another mutation.
    random.shuffle(from_vertex_ids)

    to_vertex_ids = _find_allowed_vertices(dna, self._from_regex, ...)
    if not to_vertex_ids:
      raise exceptions.MutationException()  # Try another mutation.
    random.shuffle(to_vertex_ids)

    for to_vertex_id in to_vertex_ids:
      # Avoid back-connections.
      disallowed_from_vertex_ids, _ = topology.propagated_set(to_vertex_id)
      for from_vertex_id in from_vertex_ids:
        if from_vertex_id in disallowed_from_vertex_ids:
          continue
        # This pair does not generate a cycle, so we yield it.
        yield from_vertex_id, to_vertex_id

  def _mutate_structure(self, dna, from_vertex_id, to_vertex_id):
    """Adds the edge to the DNA instance."""
    edge_id = _random_id()
    edge_type = random.choice(self._edge_types)
    if dna.has_edge(from_vertex_id, to_vertex_id):
      return False
    else:
      new_edge = dna.add_edge(from_vertex_id, to_vertex_id, edge_type, edge_id)
      ...
      return True
\end{lstlisting}
For clarity, we omitted the details of a vertex ID targeting mechanism based on regular expressions, which is used to constrain where the additional edges are placed. This mechanism ensured the skip connections only joined points in the ``main convolutional backbone'' of the convnet. The precedence range is used to give the main backbone precedence over the skip connections when resolving scale and depth conflicts in the presence of multiple incoming edges to a vertex. Also omitted are details about the attributes of the edge to add.

To evaluate an individual's fitness, its \lstinline{DNA} is unfolded into a TensorFlow model by the \lstinline{Model} class. This describes how each \lstinline{Vertex} and \lstinline{Edge} should be interpreted. For example:
\smallskip
\begin{lstlisting}
class Model(object):
  ...

  def _compute_vertex_nonlinearity(self, tensor, vertex):
    """Applies the necessary vertex operations depending on the vertex type."""
    if vertex.type == LINEAR:
      pass
    elif vertex.type == BN_RELU:
      tensor = slim.batch_norm(
          inputs=tensor, decay=0.9, center=True, scale=True,
          epsilon=self._batch_norm_epsilon,
          activation_fn=None, updates_collections=None,
          is_training=self.is_training, scope='batch_norm')
      tensor = tf.maximum(tensor, vertex.leakiness * tensor, name='relu')
    else:
      raise NotImplementedError()
    return tensor
    
  def _compute_edge_connection(self, tensor, edge, init_scale):
    """Applies the necessary edge connection ops depending on the edge type."""
    scale, depth = self._get_scale_and_depth(tensor)
    if edge.type == CONV:
      scale_out = scale
      depth_out = edge.depth_out(depth)
      stride = 2**edge.stride_scale
      # `init_scale` is used to normalize the initial weights in the case of
      # multiple incoming edges.
      weights_initializer = slim.variance_scaling_initializer(
          factor=2.0 * init_scale**2, uniform=False)
      weights_regularizer = slim.l2_regularizer(
          weight=self._dna.weight_decay_rate)
      tensor = slim.conv2d(
          inputs=tensor, num_outputs=depth_out,
          kernel_size=[edge.filter_width(), edge.filter_height()],
          stride=stride, weights_initializer=weights_initializer,
          weights_regularizer=weights_regularizer, biases_initializer=None,
          activation_fn=None, scope='conv')
    elif edge.type == IDENTITY:
      pass
    else:
      raise NotImplementedError()
    return tensor
\end{lstlisting}

The training and evaluation (Section~\ref{training_and_validation_section}) is done in a fairly standard way, similar to that in the tensorflow.org tutorials for image models. The individual's fitness is the accuracy on a held-out validation dataset, as described in the main text.

Parents are able to pass some of their learned weights to their children (Section~\ref{weight_inheritance_section}). When a child is constructed from a parent, it inherits IDs for the different sets of trainable weights (convolution filters, batch norm shifts, etc.). These IDs are embedded in the TensorFlow variable names. When the child's weights are initialized, those that have a matching ID in the parent are inherited, provided they have the same shape:
\smallskip
\begin{lstlisting}
graph = tf.Graph()
with graph.as_default():
  # Build the neural network using the `Model` class and the `DNA` instance.
  ...

  tf.Session.reset(self._master)
  with tf.Session(self._master, graph=graph) as sess:
    # Initialize all variables
    ...

    # Make sure we can inherit batch-norm variables properly.
    # The TF-slim batch-norm variables must be handled separately here because some
    # of them are not trainable (the moving averages).
    batch_norm_extras = [x for x in tf.all_variables() if (
        x.name.find('moving_var') != -1 or
        x.name.find('moving_mean') != -1)]
        
    # These are the variables that we will attempt to inherit from the parent.
    vars_to_restore = tf.trainable_variables() + batch_norm_extras
    
    # Copy as many of the weights as possible.
    if mutated_weights:
      assignments = []
      for var in vars_to_restore:
        stripped_name = var.name.split(':')[0]
        if stripped_name in mutated_weights:
          shape_mutated = mutated_weights[stripped_name].shape
          shape_needed = var.get_shape()
          if shape_mutated == shape_needed:
            assignments.append(var.assign(mutated_weights[stripped_name]))
      sess.run(assignments)
\end{lstlisting}

\section{FLOPs estimation}

This section describes how we estimate the number of floating point operations (FLOPs) required for an entire evolution experiment. To obtain the total FLOPs, we sum the FLOPs for each individual ever constructed. An individual's FLOPs are the sum of its training and validation FLOPs. Namely, the individual FLOPs are given by $F_t N_t + F_v N_v$, where $F_t$ is the FLOPs in one training step, $N_t$ is the number of training steps, $F_v$ is the FLOPs required to evaluate one validation batch of examples and $N_v$ is the number of validation batches.

The number of training steps and the number of validation batches are known in advance and are constant throughout the experiment. $F_t$ was obtained analytically as the sum of the FLOPs required to compute each operation executed during training (that is, each node in the TensorFlow graph). $F_v$ was found analogously.

Below is the code snippet that computes FLOPs for the training of one individual, for example.
\smallskip
\begin{lstlisting}
import tensorflow as tf
tfprof_logger = tf.contrib.tfprof.python.tools.tfprof.tfprof_logger
 

def compute_flops():
  """Compute flops for one iteration of training."""
  graph = tf.Graph()
  with graph.as_default():
    # Build model
    ...

    # Run one iteration of training and collect run metadata.
    # This metadata will be used to determine the nodes which were
    # actually executed as well as their argument shapes.
    run_meta = tf.RunMetadata()
    with tf.Session(graph=graph) as sess:
      feed_dict = {...}
      _ = sess.run(
        [train_op],
        feed_dict=feed_dict,
        run_metadata=run_meta,
        options=tf.RunOptions(trace_level=tf.RunOptions.FULL_TRACE))
 
    # Compute analytical FLOPs for all nodes in the graph.
    logged_ops = tfprof_logger._get_logged_ops(graph, run_meta=run_metadata)
 
    # Determine which nodes were executed during one training step
    # by looking at elapsed execution time of each node.
    elapsed_us_for_ops = {}
    for dev_stat in run_metadata.step_stats.dev_stats:
      for node_stat in dev_stat.node_stats:
        name = node_stat.node_name
        elapsed_us = node_stat.op_end_rel_micros - node_stat.op_start_rel_micros
        elapsed_us_for_ops[name] = elapsed_us
 
    # Compute FLOPs of executed nodes.
    total_flops = 0
    for op in graph.get_operations():
      name = op.name
      if elapsed_us_for_ops.get(name, 0) > 0 and name in logged_ops:
        total_flops += logged_ops[name].float_ops
    
    return total_flops
\end{lstlisting}

Note that we also need to declare how to compute FLOPs for each operation type present (that is, for each node type in the TensorFlow graph). We did this for the following operation types (and their gradients, where applicable):
\begin{itemize}
    \item unary math operations: square, squre root, log, negation, element-wise inverse, softmax,
      L2 norm;
    \item binary element-wise operations: addition, subtraction, multiplication, division, minimum,
      maximum, power, squared difference, comparison operations;
    \item reduction operations: mean, sum, argmax, argmin;
    \item convolution, average pooling, max pooling;
    \item matrix multiplication.
\end{itemize}

For example, for the element-wise addition operation type:

\smallskip
\begin{lstlisting}
from tensorflow.python.framework import graph_util
from tensorflow.python.framework import ops
 
@ops.RegisterStatistics("Add", "flops")
def _add_flops(graph, node):
  """Compute flops for the Add operation."""
  out_shape = graph_util.tensor_shape_from_node_def_name(graph, node.name)
  out_shape.assert_is_fully_defined()
  return ops.OpStats("flops", out_shape.num_elements())

\end{lstlisting}

\section{Escaping Local Optima Details}

\label{escaping_local_optima_supplementary_section}

\subsection{Local optima and mutation rate}

Entrapment at a local optimum may mean a general lack of exploration in our search algorithm. To encourage more exploration, we increased the {\em mutation rate} (Section~\ref{analysis_section}). In more detail, we carried out experiments in which we first waited until the populations converged. Some reached higher fitnesses and others got trapped at poor local optima. At this point, we modified the algorithm slightly: instead of performing 1 mutation at each reproduction event, we performed 5 mutations. We evolved with this increased mutation rate for a while and finally we switched back to the original single-mutation version. During the 5-mutation stage, some populations escape the local optimum, as in Figure~\ref{escape_figure} (top), and none get worse. Across populations, however, the escape was not frequent enough (8 out of 10) and took too long for us to propose this as an efficient technique to escape optima. An interesting direction for future work would be to study more elegant methods to manage the exploration vs. exploitation trade-off in large-scale neuro-evolution.

\subsection{Local optima and weight resetting}

The identity mutation offers a mechanism for populations to get trapped in local optima. Some individuals may get trained more than their peers just because they happen to have undergone more identity mutations. It may, therefore, occur that a poor architecture may become more accurate than potentially better architectures that still need more training. In the extreme case, the well-trained poor architecture may become a super-fit individual and take over the population. Suspecting this scenario, we performed experiments in which we simultaneously reset all the weights in a population that had plateaued (Section~\ref{analysis_section}). The simultaneous reset should put all the individuals on the same footing, so individuals that had accidentally trained more no longer have the unfair advantage. Indeed, the results matched our expectation. The populations suffer a temporary degradation in fitness immediately after the reset, as the individuals need to retrain. Later, however, the populations end up reaching higher optima (for example, Figure~\ref{escape_figure}, bottom). Across 10 experiments, we find that three successive resets tend to cause improvement ($p < 0.001$). We mention this effect merely as evidence of this particular drawback of weight inheritance. In our main results, we circumvented the problem by using longer training times and larger populations. Future work may explore more efficient solutions.

\end{document}